\documentclass[10pt,twocolumn,letterpaper]{article}

\usepackage{cvpr}
\usepackage{times}
\usepackage{epsfig}
\usepackage{graphicx}
\usepackage{amsmath}
\usepackage{amssymb}
\usepackage{float}
\usepackage{subcaption}
\usepackage{csquotes}
\usepackage{wrapfig}
\usepackage[export]{adjustbox}
\usepackage{makecell}
\usepackage{url}

\usepackage[pagebackref=true,breaklinks=true,letterpaper=true,colorlinks,bookmarks=false]{hyperref}

\cvprfinalcopy 

\def\cvprPaperID{****} 

\ifcvprfinal\fi
\begin{document}

\title{MediaPipe Hands: On-device Real-time Hand Tracking}

\author{
Fan Zhang\quad Valentin Bazarevsky\quad Andrey Vakunov\\
Andrei Tkachenka\quad George Sung\quad Chuo-Ling Chang\quad Matthias Grundmann\\
Google Research\\
1600 Amphitheatre Pkwy, Mountain View, CA 94043, USA\\
{\tt\small \{zhafang, valik, vakunov, atkach, gsung, chuoling, grundman\}@google.com}\\
}

\maketitle

\begin{abstract}
We present a real-time on-device hand tracking solution that predicts a hand skeleton of a human from a single RGB camera for AR/VR applications. Our pipeline consists of two models: 1) a palm detector, that is providing a bounding box of a hand to, 2) a hand landmark model, that is predicting the hand skeleton. It is implemented via MediaPipe\cite{mediapipe}, a framework for building cross-platform ML solutions. The proposed model and pipeline architecture demonstrate real-time inference speed on mobile GPUs with high prediction quality. MediaPipe Hands is open sourced at \small{\url{https://mediapipe.dev}}.
\end{abstract}

\section{Introduction}
Hand tracking is a vital component to provide a natural way for interaction and communication in AR/VR, and has been an active research topic in the industry \cite{oculus} \cite{snap}. Vision-based hand pose estimation has been studied for many years. A large portion of previous work requires specialized hardware, \eg depth sensors \cite{oikonomidis2011efficient}\cite{tagliasacchi2015robust}\cite{wan2019self}\cite{ge2016robust}\cite{ge2018robust}. Other solutions are not lightweight enough to run real-time on commodity mobile devices\cite{ge20193d} and thus are limited to platforms equipped with powerful processors. In this paper, we propose a novel solution that does not require any additional hardware and performs in real-time on mobile devices. Our main contributions are:
\vspace{-\topsep}
\begin{itemize}
\setlength{\parskip}{0pt}
\setlength{\itemsep}{0pt plus 1pt}
    \item An efficient two-stage hand tracking pipeline that can track multiple hands in real-time on mobile devices.
    \item A hand pose estimation model that is capable of predicting 2.5D hand pose with only RGB input.
    \item And open source hand tracking pipeline as a ready-to-go solution on a variety of platforms, including Android, iOS, Web (Tensorflow.js\cite{tfjs}) and desktop PCs.
\end{itemize}

\begin{figure}[t!] 
\begin{center}
   \includegraphics[width=1\linewidth]{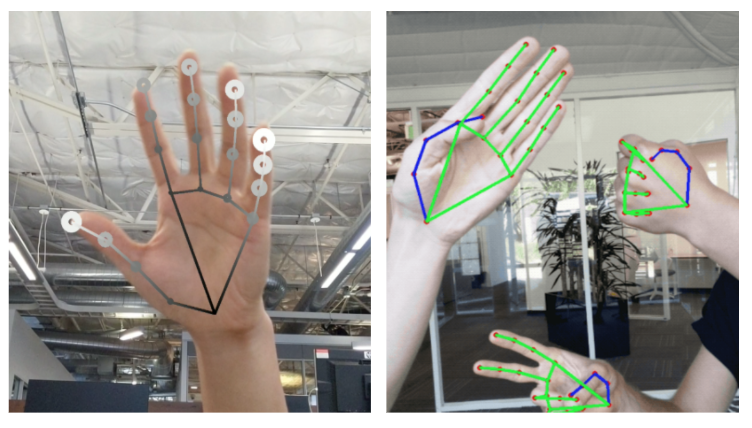}
\end{center}
   \caption{Rendered hand tracking result. (Left): Hand landmarks with relative depth presented in different shades. The lighter and larger the circle, the closer the landmark is towards the camera. (Right): Real-time multi-hand tracking on Pixel 3.}
\label{fig:hand_tracking}
\end{figure}

\section{Architecture} \label{architecture}


Our hand tracking solution utilizes an ML pipeline consisting of two models working together:
\vspace{-\topsep}
\begin{itemize}
\setlength{\parskip}{0pt}
\setlength{\itemsep}{0pt plus 1pt}
    \item A palm detector that operates on a full input image and locates palms via an oriented hand bounding box.
    \item A hand landmark model that operates on the cropped hand bounding box provided by the palm detector and returns high-fidelity 2.5D landmarks.
\end{itemize}

 Providing the accurately cropped palm image to the hand landmark model drastically reduces the need for data augmentation (\eg rotations, translation and scale) and allows the network to dedicate most of its capacity towards landmark localization accuracy. In a real-time tracking scenario, we derive a bounding box from the landmark prediction of the previous frame as input for the current frame, thus avoiding applying the detector on every frame. Instead, the detector is only applied on the first frame or when the hand prediction indicates that the hand is lost.

\subsection{BlazePalm Detector}

To detect initial hand locations, we employ a single-shot detector model optimized for mobile real-time application similar to BlazeFace\cite{blazeface}, which is also available in MediaPipe\cite{mediapipe}. Detecting hands is a decidedly complex task: our model has to work across a variety of hand sizes with a large scale span (\( \sim \)20x) and be able to detect occluded and self-occluded hands. Whereas faces have high contrast patterns, \eg, around the eye and mouth region, the lack of such features in hands makes it comparatively difficult to detect them reliably from their visual features alone.

Our solution addresses the above challenges using different strategies. 

First, we train a palm detector instead of a hand detector, since estimating bounding boxes of rigid objects like palms and fists is significantly simpler than detecting hands with articulated fingers. In addition, as palms are smaller objects, the non-maximum suppression algorithm works well even for the two-hand self-occlusion cases, like handshakes. Moreover, palms can be modelled using only square bounding boxes \cite{ssd}, ignoring other aspect ratios, and therefore reducing the number of anchors by a factor of 3\( \sim \)5. 

Second, we use an encoder-decoder feature extractor similar to FPN\cite{fpn} for a larger scene-context awareness even for small objects.

\begin{figure}[t!] 
\begin{center}
   \includegraphics[width=0.85\linewidth]{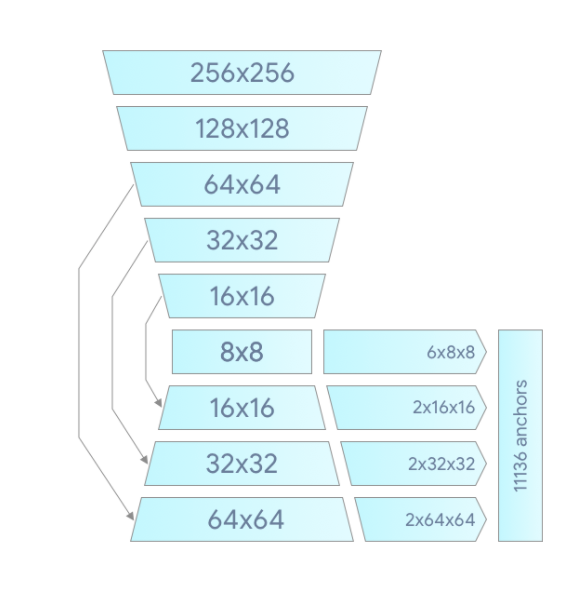}
\end{center}
   \caption{Palm detector model architecture.}
\label{fig:detectorpic}
\end{figure}

Lastly, we minimize the focal loss\cite{focalloss} during training to support a large amount of anchors resulting from the high scale variance. High-level palm detector architecture is shown in Figure \ref{fig:detectorpic}. We present an ablation study of our design elements in Table \ref{tab:detector}.

\subsection{Hand Landmark Model} \label{hand_landmark_model}

\begin{figure}[t!] 
\begin{center}
   \includegraphics[width=1.0\linewidth]{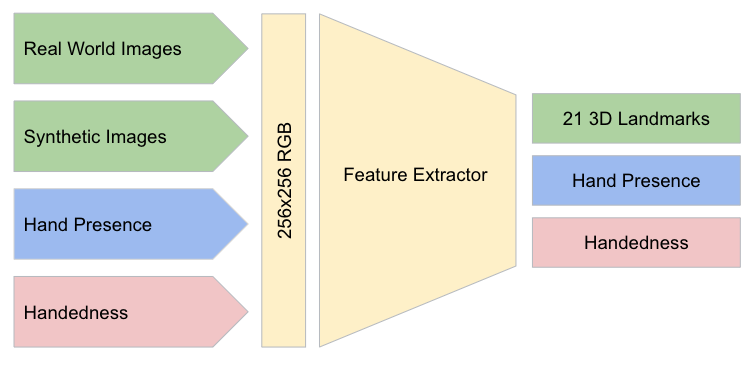}
\end{center}
   \caption{Architecture of our hand landmark model. The model has three outputs sharing a feature extractor. Each head is trained by correspondent datasets marked in the same color. See Section \ref{hand_landmark_model} for more detail.}
\label{fig:landmarkpic}
\end{figure}

After running palm detection over the whole image, our subsequent hand landmark model performs precise landmark localization of 21 2.5D coordinates inside the detected hand regions via regression. The model learns a consistent internal hand pose representation and is robust even to partially visible hands and self-occlusions. The model has three outputs (see Figure \ref{fig:landmarkpic}):
\vspace{-\topsep}
\begin{enumerate}
\setlength{\parskip}{0pt}
\setlength{\itemsep}{0pt plus 1pt}
    \item 21 hand landmarks consisting of x, y, and relative depth.
    \item A hand flag indicating the probability of hand presence in the input image.
    \item A binary classification of handedness, \eg left or right hand. 
\end{enumerate}
\vspace{-\topsep}

We use the same topology as \cite{cmuhand} for the 21 landmarks. The 2D coordinates are learned from both real-world images as well as synthetic datasets as discussed below, with the relative depth \wrt the wrist point being learned only from synthetic images. To recover from tracking failure, we developed another output of the model similar to \cite{facemesh} for producing the probability of the event that a reasonably aligned hand is indeed present in the provided crop. If the score is lower than a threshold then the detector is triggered to reset tracking. Handedness is another important attribute for effective interaction using hands in AR/VR. This is especially useful for some applications where each hand is associated with a unique functionality. Thus we developed a binary classification head to predict whether the input hand is the left or right hand.
Our setup targets real-time mobile GPU inference, but we have also designed lighter and heavier versions of the model to address CPU inference on the mobile devices lacking proper GPU support and higher accuracy requirements of accuracy to run on desktop, respectively.

\section{Dataset and Annotation} \label{dataset}

\begin{figure}[t!] 
\begin{center}
   \includegraphics[width=1.0\linewidth]{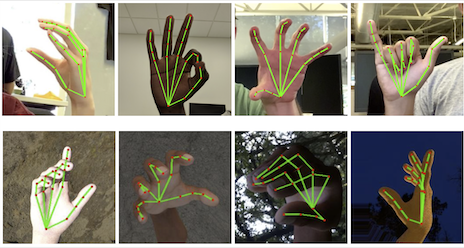}
\end{center}
   \caption{Examples of our datasets. (Top): Annotated real-world images. (Bottom): Rendered synthetic hand images with ground truth annotation. See Section \ref{dataset} for details.}
\label{fig:dataset}
\end{figure}

To obtain ground truth data, we created the following datasets addressing different aspects of the problem:

\vspace{-\topsep}
\begin{itemize}
\setlength{\parskip}{0pt}
\setlength{\itemsep}{0pt plus 1pt}
    \item In-the-wild dataset: This dataset contains 6K images of large variety, \eg geographical diversity, various lighting conditions and hand appearance. The limitation of this dataset is that it doesn't contain complex articulation of hands.
    \item In-house collected gesture dataset: This dataset contains 10K images that cover various angles of all physically possible hand gestures. The limitation of this dataset is that it's collected from only 30 people with limited variation in background. The in-the-wild and in-house dataset are great complements to each other to improve robustness.
    \item Synthetic dataset: To even better cover the possible hand poses and provide additional supervision for depth, we render a high-quality synthetic hand model over various backgrounds and map it to the corresponding 3D coordinates. We use a commercial 3D hand model that is rigged with 24 bones and includes 36 blendshapes, which control fingers and palm thickness. The model also provides 5 textures with different skin tones. We created video sequences of transformation between hand poses and sampled 100K images from the videos. We rendered each pose with a random high-dynamic-range lighting environment and three different cameras. See Figure \ref{fig:dataset} for examples.
\end{itemize}

For the palm detector, we only use in-the-wild dataset, which is sufficient for localizing hands and offers the highest variety in appearance. However, all datasets are used for training the hand landmark model. We annotate the real-world images with 21 landmarks and use projected ground-truth 3D joints for synthetic images. For hand presence, we select a subset of real-world images as positive examples and sample on the region excluding annotated hand regions as negative examples. For handedness, we annotate a subset of real-world images with handedness to provide such data.

\section{Results} \label{results}

For the hand landmark model, our experiments show that the combination of real-world and synthetic datasets provides the best results. See Table \ref{tab:datasets} for details. We evaluate only on real-world images. Beyond the quality improvement, training with a large synthetic dataset leads to less jitter visually across frames. This observation leads us to believe that our real-world dataset can be enlarged for better generalization. 

\begin{table}[h!]
  \begin{center}
    \begin{tabular}{c|c} 
      \textbf{Model Variation} & \textbf{Average Precision}\\
      \hline
      No decoder + cross entropy loss & 86.22\%\\
      Decoder + cross entropy loss & 94.07\%\\
      Decoder + focal loss & 95.7\%\\
    \end{tabular}
  \caption{Ablation study of palm detector design elements of palm detector.}
  \label{tab:detector}
  \end{center}
\end{table}

\begin{table}[h!]
  \begin{center}
    \begin{tabular}{l|c} 
      \textbf{Dataset} & \textbf{MSE normalized by palm size}\\
      \hline
      Only real-world & 16.1\%\\
      Only synthetic & 25.7\%\\
      Combined & 13.4\%\\
    \end{tabular}
  \caption{Results of our model trained from different datasets.}
  \label{tab:datasets}
  \end{center}
\end{table}

Our target is to achieve real-time performance on mobile devices. We experimented with different model sizes and found that the \enquote{Full} model (see Table \ref{tab:performance}) provides a good trade-off between quality and speed. Increasing model capacity further introduces only minor improvements in quality but decreases significantly in speed (see Table \ref{tab:performance} for details). We use the TensorFlow Lite GPU backend for on-device inference\cite{tflite}.

\begin{table}[h!]
  \begin{center}
    \begin{tabular}{l|c|c|c|c|c} 
      \thead{\textbf{Model}} & \thead{\textbf{Params} \\ \textbf{(M)}} & \thead{\textbf{MSE}} & \thead{\textbf{Time(ms)} \\ \textbf{Pixel 3}} & \thead{\textbf{Time(ms)} \\ \textbf{Samsung} \\ \textbf{S20}} & \thead{\textbf{Time(ms)} \\ \textbf{iPhone11}}\\
      \hline
      Light & 1 & 11.83 & 6.6 & 5.6 & 1.1\\
      Full & 1.98 & 10.05 & 16.1 & 11.1 & 5.3\\
      Heavy & 4.02 & 9.817 & 36.9 & 25.8 & 7.5\\
    \end{tabular}
  \caption{Hand landmark model performance characteristics.}
  \label{tab:performance}
  \end{center}
\end{table}

\section{Implementation in MediaPipe} \label{implementation}

With MediaPipe\cite{mediapipe}, our hand tracking pipeline can be built as a directed graph of modular components, called Calculators. Mediapipe comes with an extensible set of Calculators to solve tasks like model inference, media processing, and data transformations across a wide variety of devices and platforms. Individual Calculators like cropping, rendering and neural network computations are further optimized to utilize GPU acceleration. For example, we employ TFLite GPU inference on most modern phones.

\begin{figure}[t!] 
\begin{center}
   \includegraphics[width=1.0\linewidth]{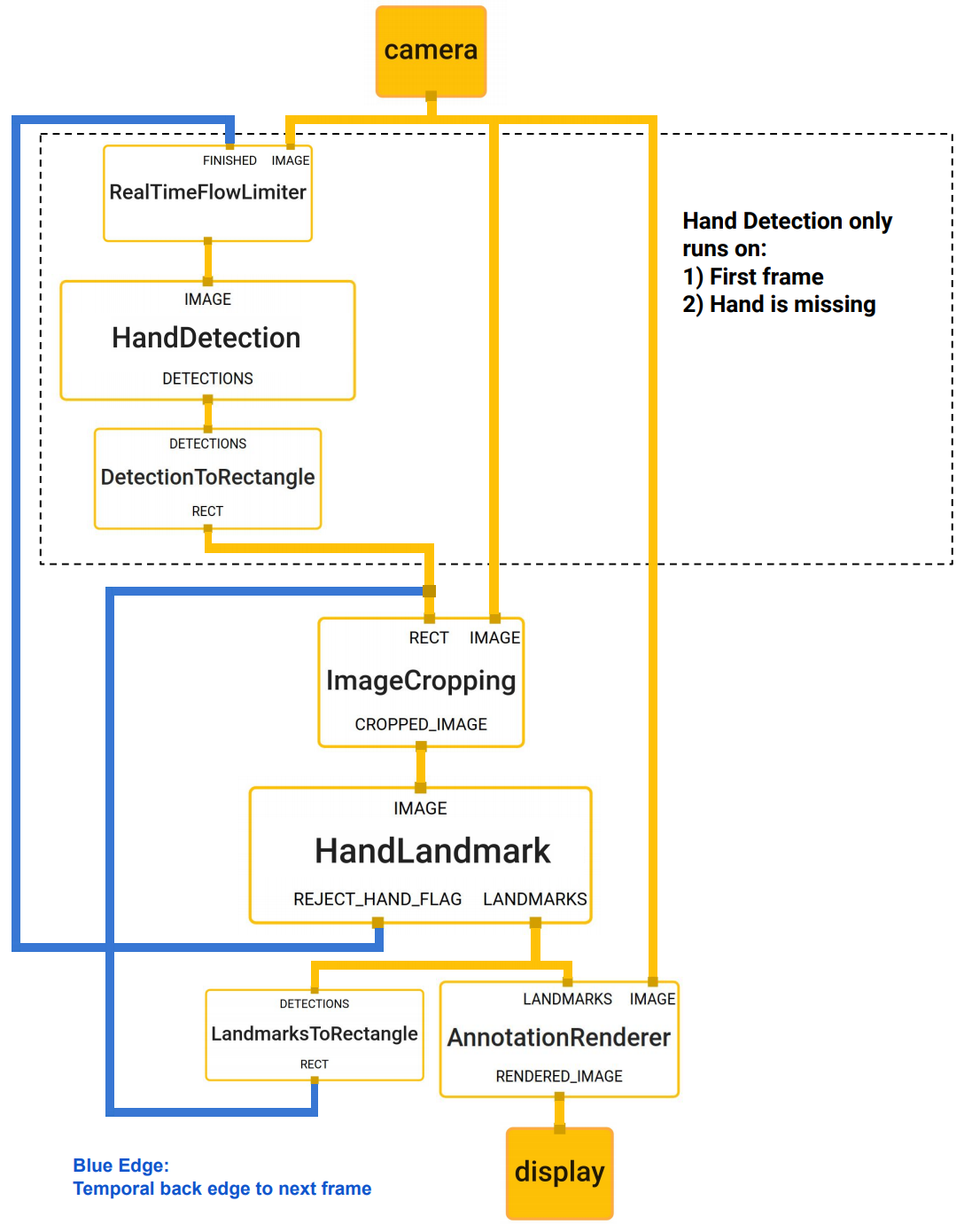}
\end{center}
   \caption{The hand landmark model’s output controls when the hand detection model is triggered. This behavior is achieved by MediaPipe’s powerful synchronization building blocks, resulting in high performance and optimal throughput of the ML pipeline.}
\label{fig:mediapipe_graph}
\end{figure}

Our MediaPipe graph for hand tracking is shown in Figure \ref{fig:mediapipe_graph}. The graph consists of two subgraphs — one for hand detection and another for landmarks computation. One key optimization MediaPipe provides is that the palm detector only runs as needed (fairly infrequently), saving significant computation. We achieve this by deriving the hand location in the current video frames from the computed hand landmarks in the previous frame, eliminating the need to apply the palm detector on every frame. For robustness, the hand tracker model also outputs an additional scalar capturing the confidence that a hand is present and reasonably aligned in the input crop. Only when the confidence falls below a certain threshold is the hand detection model reapplied to the next frame.

\section{Application examples} \label{application}

\begin{figure}[t!] 
\begin{center}
   \includegraphics[width=1.0\linewidth]{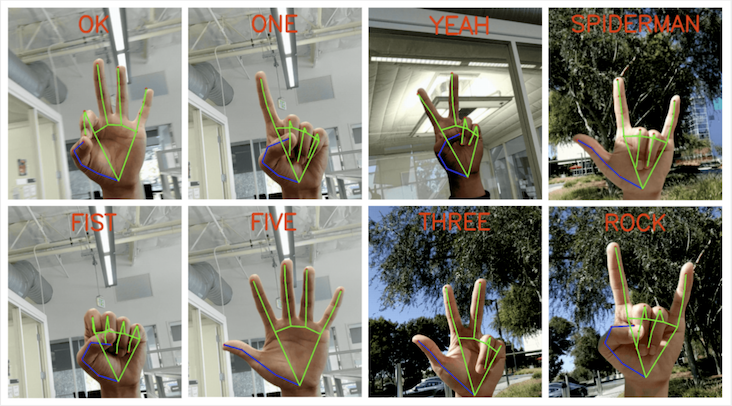}
\end{center}
   \caption{Screenshots of real-time gesture recognition. Semantics of gestures are rendered at top of the images.}
\label{fig:gesture}
\end{figure}

\begin{figure}[t!] 
\begin{center}
   \includegraphics[width=1.0\linewidth]{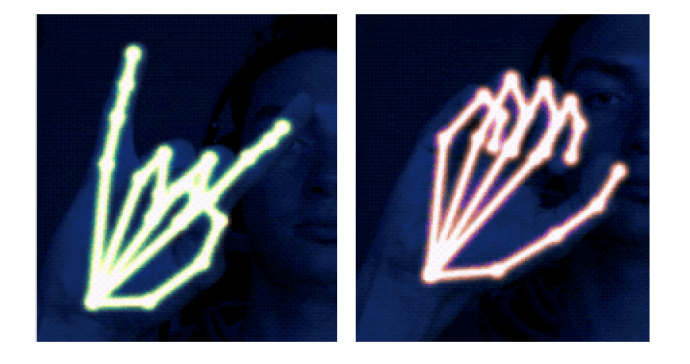}
\end{center}
   \caption{Example of real-time AR effects based on our predicted hand skeleton.}
\label{fig:areffect}
\end{figure}

Our hand tracking solution can readily be used in many applications such as gesture recognition and AR effects. On top of the predicted hand skeleton, we employ a simple algorithm to compute gestures, see Figure \ref{fig:gesture}. First, the state of each finger, \eg bent or straight, is determined via the accumulated angles of joints. Then, we map the set of finger states to a set of predefined gestures. This straightforward, yet effective technique allows us to estimate basic static gestures with reasonable quality. Beyond static gesture recognition, it is also possible to use a sequence of landmarks to predict dynamic gestures. Another application is to apply AR effects on top of the skeleton. Hand based AR effects currently enjoy high popularity. In Figure \ref{fig:areffect}, we show an example AR rendering of the hand skeleton in neon light style.

\section{Conclusion}

In this paper, we proposed MediaPipe Hands, an end-to-end hand tracking solution that achieves real-time performance on multiple platforms. Our pipeline predicts 2.5D landmarks without any specialized hardware and thus, can be easily deployed to commodity devices. We open sourced the pipeline to encourage researchers and engineers to build gesture control and creative AR/VR applications with our pipeline.

{\small
\bibliographystyle{ieee_fullname}
\bibliography{egbib}
}

\end{document}